\documentclass[journal,twoside,web]{ieeecolor}
\usepackage{generic}
\usepackage{cite}
\usepackage{amsmath,amssymb,amsfonts}
\usepackage{algorithmic}
\usepackage{graphicx}
\usepackage{algorithm,algorithmic}
\usepackage{hyperref}

\usepackage{verbatim}
\usepackage{booktabs}
\usepackage{multirow}
\usepackage{stfloats}
\usepackage{marvosym}
\usepackage{caption}
\usepackage{tabularx}
\usepackage{array}
\usepackage{pifont}
\usepackage{xcolor}
\usepackage[table,dvipsnames,x11names]{xcolor}
\usepackage{orcidlink}

\newcommand{\cmark}{\ding{51}}

\newcommand{\ie}{\textit{i}.\textit{e}.}
\newcommand{\eg}{\textit{e}.\textit{g}.}

\hypersetup{hidelinks=true}
\usepackage{textcomp}
\def\BibTeX{{\rm B\kern-.05em{\sc i\kern-.025em b}\kern-.08em
    T\kern-.1667em\lower.7ex\hbox{E}\kern-.125emX}}

\usepackage{etoolbox}
\usepackage{fancyhdr}
\fancypagestyle{plain}{%
  \fancyhf{}

}

\makeatletter
\gdef\@preprint{}

\gdef\@authorhead{}  

\def\ps@pprintTitle{%
  \let\@oddhead\@empty\let\@evenhead\@empty
  \let\@oddfoot\@empty\let\@evenfoot\@empty
}

\def\ps@title{%
  \let\@oddhead\@empty\let\@evenhead\@empty
  \let\@oddfoot\@empty\let\@evenfoot\@empty
}

\let\old@maketitle\maketitle
\renewcommand{\maketitle}{%
  \gdef\@preprint{}%
  \gdef\@authorhead{}%
  \old@maketitle
  \thispagestyle{fancy}%
}
\makeatother
\begin{document}
\title{DECIS: Dual-Evidence Corrective Verification for Interpretable Strabismus Diagnostic Decision-Making}
\author{Xikai Tang\orcidlink{0000-0002-5231-8200}, Yifan Wang, Jiafan Zhuang\orcidlink{0000-0003-3708-4634}, Li Luo, Jinming Guo, Xiaoling Xie, Jiacheng Liu, Peiwei Wei, Lihao Zhong, Xiaoli Kang, Jie Cen, Guangqiang Yin, Kunliang Qiu, Ce Zheng\orcidlink{0000-0001-7146-9138}, and Zhun Fan\orcidlink{0000-0002-4232-8229},~\IEEEmembership{Senior Member,~IEEE}
\thanks{This work was supported in part by the National Natural Science Foundation of China (grant numbers 62406186, 62476163, and 82571270), the Hospital Funded Clinical Research, Xinhua Hospital Affiliated to Shanghai Jiao Tong University School of Medicine (grant numbers 21XJMR02 and 24XHCR10B), the Natural Science Foundation of Guangdong Province (grant number 2025A1515010800), the Guangdong Basic and Applied Basic Research Foundation (grant number 2023B1515120020), the GBA Ascend Application Innovation Institute, and the Guangdong Laboratory of Artificial Intelligence and Digital Economy (SZ) (grant number GML-ST-2026-02), the Industry-University-Research Innovation Fund of Chinese Universities--New Generation Information Technology Innovation Project (grant number 2024IT013), the Institutional Project of Joint Shantou International Eye Center of Shantou University and the Chinese University of Hong Kong (grant number 26-008), the State Key Laboratory of Autonomous Intelligent Unmanned Systems (grant number ZZKF2025-3-4).}
\thanks{
Xikai Tang and Guangqiang Yin are with the School of Information and
Software Engineering, University of Electronic Science and Technology
of China, Chengdu 611731, China.
}
\thanks{
Yifan Wang, Jiafan Zhuang, and Zhun Fan are with the Shenzhen Institute
for Advanced Study, University of Electronic Science and Technology of
China, Shenzhen 518000, China.
}
\thanks{
Li Luo, Jinming Guo, Xiaoling Xie, and Kunliang Qiu are with the Joint
Shantou International Eye Center of Shantou University and The Chinese
University of Hong Kong, Shantou 515041, China.
}
\thanks{
Jiacheng Liu is with the School of Artificial Intelligence, Guangzhou
City Polytechnic, Guangzhou 510405, China.
}
\thanks{
Peiwei Wei is with the Medical College, Shantou University,
Shantou 515041, China.
}
\thanks{
Lihao Zhong is with the College of Engineering, Shantou University,
Shantou 515063, China.
}
\thanks{
Xiaoli Kang, Jie Cen, and Ce Zheng are with the Department of
Ophthalmology, Xinhua Hospital Affiliated to Shanghai Jiao Tong
University School of Medicine, Shanghai 200092, China.
}
\thanks{
Zhun Fan is also with the Shenzhen Loop Area Institute,
Shenzhen 518048, China.
}
%
\thanks{\textsuperscript{\Letter} Corresponding authors: Jiafan Zhuang, Kunliang Qiu, Ce Zheng and Zhun Fan}
}


\maketitle

\begin{abstract}
Strabismus is a common ocular disorder that requires fine-grained subtype diagnosis for individualized treatment planning. However, existing deep learning methods mainly provide diagnostic predictions without transparent reasoning, while recent large vision-language models (LVLMs), although promising for joint image understanding and report generation, remain highly prone to hallucination in this evidence-sensitive and rule-driven medical task. To address these challenges, we propose DECIS, a \textbf{D}ual-\textbf{E}vidence \textbf{C}orrective verification for \textbf{I}nterpretable \textbf{S}trabismus diagnostic decision-making framework. DECIS transforms black-box end-to-end generation into a structured diagnostic process consisting of candidate hypothesis generation, dual-evidence constrained context, evidence-based corrective verification, and report generation. 
Specifically, we introduce a Dual-Evidence Constrained Context (DECC)  mechanism that jointly organizes visual evidence from the photograph of the nine cardinal positions of gaze and evidence-based clinical diagnostic rules into a constrained context for reliable diagnostic reasoning.
We further develop an Evidence-Based Corrective Verification (EBCV) mechanism that verifies whether the current diagnostic hypothesis is supported by visual evidence, heatmap-based visual cues, and evidence-based clinical diagnostic rules. Hypothesis refinement is triggered when inconsistency is detected. Experiments on a fine-grained strabismus benchmark demonstrate that DECIS not only significantly outperforms other state-of-the-art diagnostic systems, improving the weighted F1 score from \textbf{72.0\%} to \textbf{91.3\%}, but also substantially improves the clinical reliability (consistency, alignment, and completeness) of generated diagnostic reports. These results demonstrate that DECIS provides an effective solution for building accurate, evidence-based, and clinically interpretable strabismus decision making systems.
\end{abstract}


\section{Introduction}
\label{sec:introduction}
Strabismus, defined as any form of binocular misalignment \cite{strabismus1}, is a common ocular disorder with a prevalence ranging from 0.8\% to 6.0\% \cite{strabismus2},\cite{strabismus3},\cite{strabismus4},\cite{strabismus5}, traditionally considered to primarily affect children \cite{strabismus6}. Accurate subtype diagnosis is essential for individualized treatment planning and surgical decision-making.
Rather than simple ocular misalignment detection, strabismus subtype diagnosis requires analyzing the deviation direction and angle variation of ocular deviation across the photograph of the nine cardinal positions of gaze \cite{Menon2002StudyOT},\cite{2011prevalence}. 
To identify specific strabismus subtypes, clinicians interpret these ocular observations according to subtype-specific diagnostic rules.
Therefore, fine-grained strabismus diagnosis is inherently an evidence-sensitive and rule-dependent task, where reliable conclusions must be supported by visual evidence and interpreted within a clinical interpretive framework.

Existing deep learning methods \cite{traditionaleye1},\cite{traditionaleye2} for strabismus analysis mainly aim to predict diagnostic labels from ocular images. Despite their encouraging performance, these methods usually operate as black-box predictors, fail to elaborate the visual 
evidence-based reasoning process of diagnostic and surgical 
decision-making. Recently, CI-GNN \cite{zheng2025causality} took an initial step toward interpretable strabismus diagnosis by identifying clinically relevant features and highlighting contributory variables. However, it still fails to establish a transparent and verifiable diagnostic process. As a result, the outputs of existing methods remain difficult to be verified according to clinical evidences, since the relationship between the predicted results and the underlying diagnostic basis is not explicitly established. This lack of process-level interpretability remains a major barrier to the clinical adoption of automated strabismus decision making systems.

\begin{figure*}
  \includegraphics[width=\textwidth]{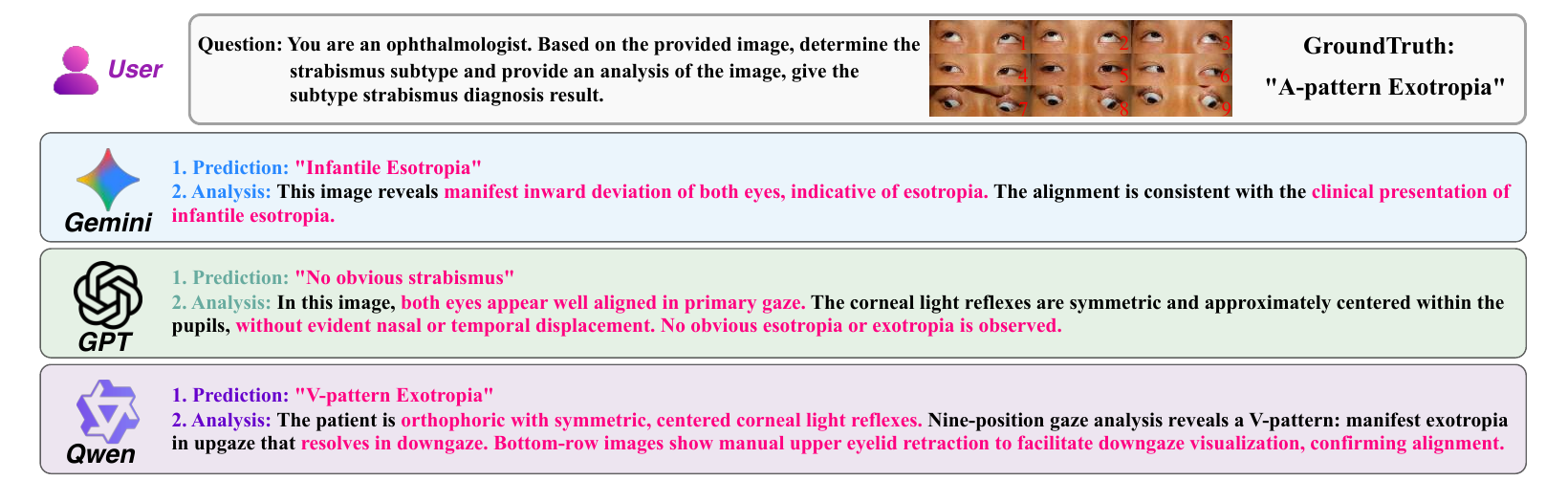}
  \caption{Visualization of strabismus diagnostic results generated by Gemini-3-Flash-Preview, GPT-5.2, and Qwen3-VL-Plus. These examples show hallucinated responses, including incorrect subtype predictions and clinically unsupported outputs. Text highlighted in red denotes hallucinated responses.}
  \label{ggq}
\end{figure*}

\begin{figure}[]
  \centering
  \includegraphics[width=\columnwidth]{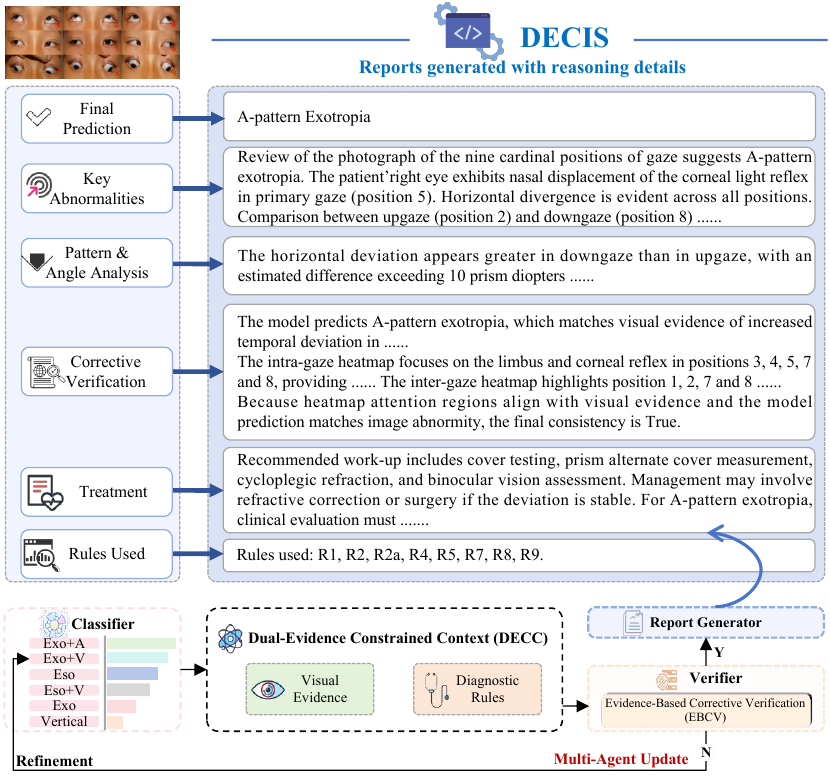}
  \caption{Overview of DECIS for strabismus subtype diagnosis. The framework uses three interacting agents and integrates Dual-Evidence Constrained Context (DECC) with Evidence-Based Corrective Verification (EBCV) to verify the classifier’s predicted subtype against visual evidence and diagnostic rules, enabling iterative refinement before report generation.}
  \label{magic}
\end{figure}

Large vision-language models (LVLMs) \cite{google2026gemini3flashpreview},\cite{2024llama},\cite{bai2025qwen3},\cite{openai2025gpt52systemcard},\cite{zhang2024disease} offer a promising alternative, as they can unify image understanding, reasoning, and diagnostic report generation within a unified framework \cite{wang2022automated},\cite{li2024ultrasound},\cite{chen2025large}. Nevertheless, they remain highly prone to hallucination in strabismus diagnosis, often producing conclusions that are inconsistent with visual evidence or misaligned with accepted clinical criteria \cite{2025survey},\cite{2024hallucination}. To evaluate this issue, we benchmarked several mainstream LVLMs, including Gemini-3-Flash-Preview \cite{google2026gemini3flashpreview}, GPT-5.2 \cite{openai2025gpt52systemcard}, and Qwen3-VL-Plus \cite{bai2025qwen3}, on strabismus subtype classification and diagnostic reasoning. As shown in Fig.~\ref{ggq}, these models produce two representative types of hallucinated responses: incorrect classification and clinically unsupported outputs. Thus, even though LVLMs may alleviate the “black-box” issue of conventional deep learning models, they also raise a critical challenge: how to ensure that the multimodal diagnostic reasoning process is reliably grounded by visual evidence and remains clinically acceptable.

Existing hallucination mitigation approaches for LVLMs can be broadly grouped into training-stage alignment \cite{zhang2024self},\cite{sun2024aligning},\cite{yu2024rlhf},\cite{liu2024multi} inference-time grounding \cite{lewis2020retrieval},\cite{JBHI3},\cite{JBHI4},\cite{JBHI5},\cite{asai2023self},\cite{2025visualmigatehall},\cite{jin2025chain},\cite{zhao2024topicwise},\cite{yang2025spatio} and post-hoc verification \cite{manakul2023selfcheckgpt},\cite{dhuliawala2024chain}. Although these strategies can reduce unreliable generation, they remain inadequate for strabismus diagnosis. A central limitation is that these methods mainly treat hallucination mitigation by controlling response-level generation, rather than using evidence-based diagnostic verification. They do not explicitly model diagnosis as a structured process in which a diagnostic hypothesis is first formulated, then examined against visual evidence in the photograph of the nine cardinal positions of gaze, and finally evaluated according to accepted clinical criteria. As a result, current methods can hardly articulate how the diagnostic conclusion is reached through a clinically verifiable process, or validate whether the results are supported by the underlying evidence.

To address these limitations, we propose \textbf{DECIS}, a \textbf{D}ual-\textbf{E}vidence \textbf{C}orrective verification for \textbf{I}nterpretable \textbf{S}trabismus diagnostic decision-making framework that reformulates LVLM-assisted diagnosis as an evidence-constrained and verifiable process as shown in Fig.~\ref{magic}. Instead of generating the final conclusion in a single pass, DECIS decomposes diagnosis into four successive stages: diagnostic hypothesis generation, Dual-Evidence Constrained Context (DECC), Evidence-Based Corrective Verification (EBCV), and report generation. 
Specifically, DECC constructs a diagnostically constrained context by combining visual evidence with relevant evidence-based clinical diagnostic rules formalized through a doctor-in-the-loop process. 
EBCV then examines whether the current diagnostic hypothesis is supported by visual evidence and evidence-based clinical diagnostic rules, and triggers hypothesis revision when inconsistency is detected. In this way, DECIS transforms strabismus diagnosis from black-box end-to-end generation into an explicit process of hypothesis formulation, evidence examination, rule-based judgment, and corrective refinement, which improves both diagnostic accuracy and interpretability.

The main contributions of this work are threefold:
\begin{itemize}
\item We propose DECIS, an evidence-based multi-agent reasoning framework that converts LVLM-assisted strabismus diagnosis from black-box generation into an evidence-constrained and verifiable diagnostic decision-making process.
\item We develop DECC to construct a dual-evidence context based on visual evidence and evidence-based clinical diagnostic rules, and EBCV to assess whether the candidate diagnosis is consistent with this context and refine inconsistent predictions.
\item We validate DECIS on a strabismus subtype benchmark and demonstrate that 
DECIS not only significantly outperforms other state-of-the-art diagnostic systems, improving the weighted F1 score from \textbf{72.0\%} to \textbf{91.3\%}, but also substantially improves the clinical reliability (clinical consistency, visual alignment, contextual completeness) of generated diagnostic reports.

\end{itemize}

\section{Related Works}

\subsection{Hallucination Mitigation in LVLMs}
Hallucination mitigation methods for LVLMs can be broadly categorized into three groups: training-stage alignment, inference-time grounding, and post-hoc verification. These methods aim to enhance the factual and evidential reliability of model outputs at different stages of the generation pipeline, but they differ substantially in how external evidence is incorporated and how hallucination is corrected.

\textbf{Training-stage alignment methods} improve factuality by injecting additional supervision into adjusting model parameters during training or fine-tuning. Representative approaches include self-alignment \cite{zhang2024self} and RLHF-based methods \cite{sun2024aligning},\cite{yu2024rlhf}, which encourage the model to better match human feedback or reference outputs. While such methods can improve general reliability, they typically require expensive retraining and do not ensure that the diagnostic conclusion for a specific case is grounded in the corresponding image findings.

\textbf{Inference-time grounding methods} constrain generation without changing model parameters by introducing external evidence during inference. Representative directions include retrieval-augmented generation \cite{lewis2020retrieval},\cite{JBHI3},\cite{JBHI4},\cite{JBHI5}, self-retrieval strategies \cite{asai2023self}, clinically guided prompting \cite{jiang2025comt},\cite{leng2024mitigating},\cite{2025visualmigatehall},\cite{jin2025chain},\cite{favero2024multi}. These approaches improve response quality by narrowing the generation space, but they still mainly operate as response-level guidance and rarely model diagnosis as an explicit and process of hypothesis formulation, evidence examination, rule-based judgment, and report generation.

\textbf{Post-hoc verification methods} attempt to detect and/or revise hallucinated outputs after generation. Typical examples include self-consistency \cite{wang2022self}, SelfCheckGPT \cite{manakul2023selfcheckgpt}, Chain-of-Verification \cite{dhuliawala2024chain}, and NLI-based (Natural Language Inference) hallucination detection \cite{chen2025explainable}. Although such methods can identify unreliable responses to some extent, they usually operate at the textual level and therefore remain insufficient for evidence-sensitive medical diagnosis, where clinically valid conclusions must be directly supported by visual evidence and accepted clinical criteria.
Overall, existing hallucination mitigation methods mainly focus on constraining the generated response. In contrast, our work formulates strabismus diagnosis as an evidence-based medical decision process, in which diagnostic conclusions are systematically supported by visual evidence and clinically relevant rules.

\subsection{Multi-Agent Collaboration for Reliable Reasoning}
Multi-agent collaboration has recently emerged as a promising paradigm for improving reasoning reliability by assigning different reasoning roles to multiple interacting agents. Compared with monolithic inference \cite{du2024improving},\cite{JBHI1},\cite{JBHI2},\cite{liang2024encouraging},\cite{chan2023chateval}, multi-agent systems can enhance reasoning depth, factual consistency, and robustness through interaction. Existing methods can be broadly categorized into two groups: cooperation/competition and reflection/verification methods.

\textbf{Cooperation/Competition methods} 
improve output quality through debate, comparison, or evaluator-based interaction among agents. Representative methods include multi-agent debate \cite{du2024improving}, automated evaluation framework\cite{JBHI1}, MedSegAgent\cite{JBHI2}, MAD \cite{liang2024encouraging} and ChatEval \cite{chan2023chateval}, in which different agents generate alternative arguments or assessments before a final response is selected or synthesized \cite{qian2024scaling}.

\textbf{Reflection/Verification methods} 
focus on iterative refinement through self-critique, self-correction, or internal consistency checking. Representative approaches include Reflexion \cite{shinn2023reflexion}, Self-Refine \cite{madaan2023self}, and Self-Verification \cite{weng2023large}, which revise intermediate reasoning steps or final outputs based on model-generated feedback \cite{ki2025multiple}.

Despite their promise, existing multi-agent methods remain insufficient for strabismus diagnosis. A key limitation is that agent interactions are still mainly driven by textual reasoning traces, with limited grounding in visual evidence and clinically relevant rules. For an evidence-sensitive medical task, this makes it difficult to verify whether the final consensus is supported by case-specific visual evidence and accepted clinical criteria. In contrast, our framework integrates multi-agent collaboration with visual evidence and clinically relevant rules, allowing candidate predictions to be explicitly verified and refined before the final diagnosis is generated.

\section{Method}
\begin{figure*}
\centerline{
\includegraphics[width=\textwidth]{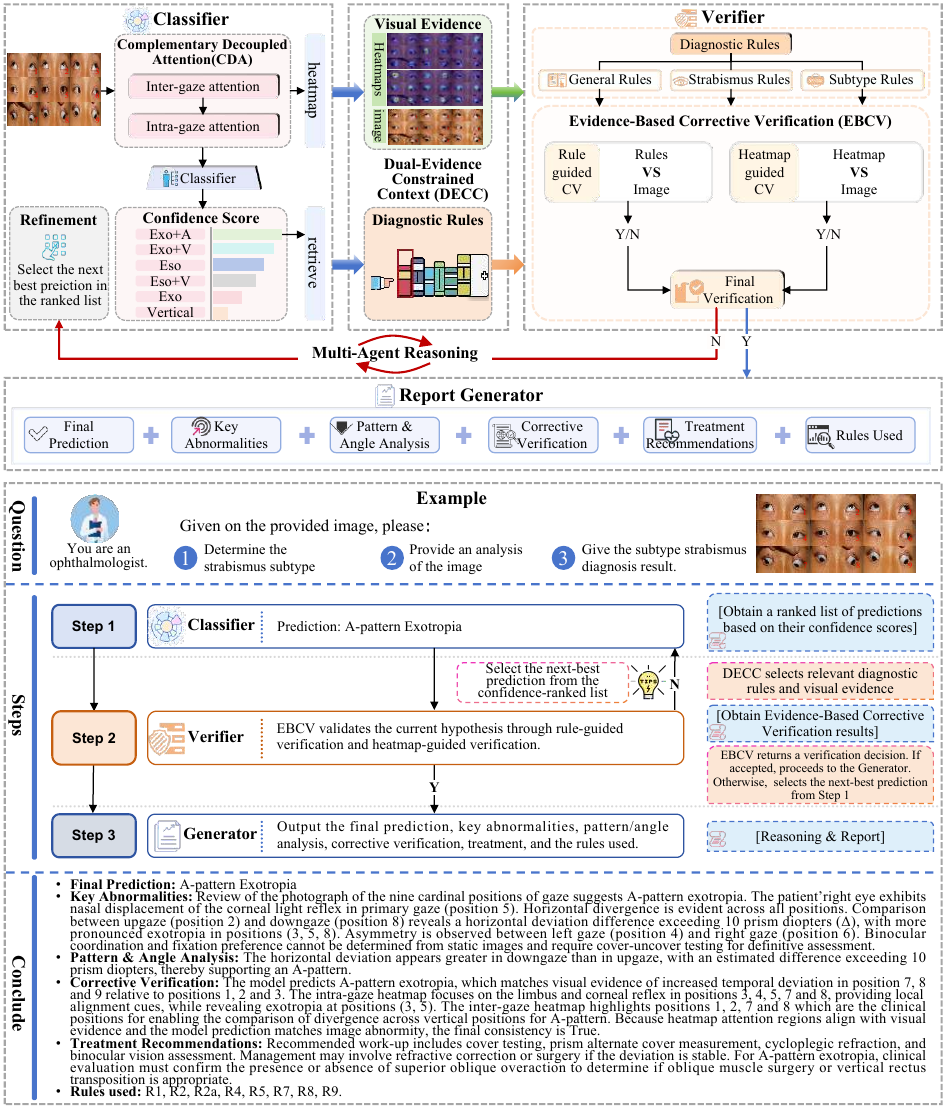}}
\caption{Overview of the DECIS framework for hallucination-mitigated strabismus subtype diagnosis. The upper panel presents the overall workflow of DECIS, which consists of a cyclic three-agent architecture integrating Dual-Evidence Constrained Context (DECC) and Evidence-Based Corrective Verification (EBCV). The Classifier first generates a confidence-ranked list of candidate subtype predictions from the input image. DECC then organizes visual evidence and formalized diagnostic rules into a diagnostically constrained context. In the verification stage, the Verifier uses EBCV to assess whether the current best candidate prediction is consistent with this context. If the current prediction is rejected, the next-best candidate from the confidence-ranked list is selected for verification. Once a candidate prediction is accepted, the Generator produces a detailed diagnostic report. The lower panel illustrates a representative case processed by DECIS.}  
\label{magic_main}
\end{figure*}

\begin{figure}
\includegraphics[width=0.5\textwidth]{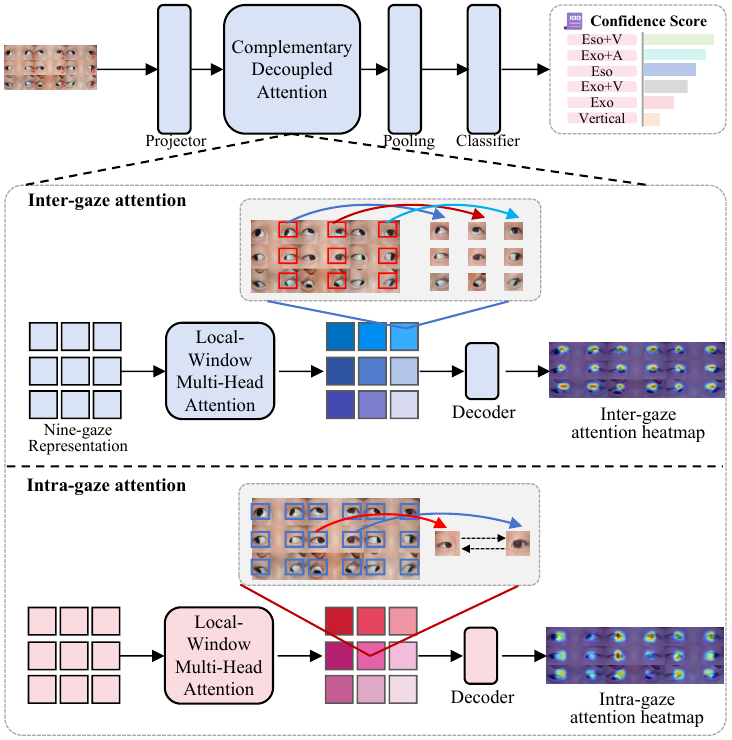}
\caption{Illustration of the Complementary Decoupled Attention (CDA) module. CDA separately models intra-gaze and inter-gaze dependencies from the photograph of the nine cardinal positions of gaze, generating attention heatmaps as visual evidence and a confidence-ranked list of candidate subtype predictions}
\label{GazeNet}
\end{figure}
\subsection{Overview}
This work proposes DECIS, a novel multi-agent system with a dual-evidence corrective verification framework enabling interpretable strabismus subtype diagnosis, personalized treatment planning, and surgical diagnostic decision-making. Unlike standard LVLM prompting pipelines that generate diagnostic conclusions in a single pass, DECIS formulates diagnosis as a process consisting of candidate hypothesis generation, dual-evidence constrained context construction, evidence-based corrective verification, and structured report generation. This design is motivated by the clinical principle that a diagnostic conclusion should not be accepted solely because it appears linguistically plausible.Instead, it must be explicitly validated against visual evidence and evidence-based clinical diagnostic rules prior to final decision-making.

The overall framework is shown in Fig.~\ref{magic_main}. DECIS is built upon three interacting agents: a Classifier Agent, a Verifier Agent, and a Generator Agent. The Classifier Agent first processes the photograph of the nine cardinal positions of gaze and produces a ranked list of diagnostic hypotheses together with visual evidence, including visual cues within each gaze positions and relational cues across gaze positions. Based on these outputs, the framework constructs a Dual-Evidence Constrained Context (DECC) by combining visual evidence with evidence-based clinical diagnostic rules formalized through a doctor-in-the-loop process. The resulting context is then passed to the Verifier Agent, which performs Evidence-Based Corrective Verification (EBCV) to determine whether the current diagnostic hypothesis is supported by visual evidence and evidence-based clinical diagnostic rules. If the hypothesis passes verification, the Generator Agent organizes the verified visual evidence into a structured diagnostic report. Otherwise, the framework triggers refinement the next best candidate hypothesis in the ranked list under updated evidence constraints.

The key idea of DECIS is to convert LVLM-assisted diagnosis from black-box end-to-end generation into an explicit and verifiable decision process. In this framework, DECC provides the constrained diagnostic context needed for grounded reasoning, while EBCV ensures that candidate conclusions are not directly accepted unless they are supported by both visual evidence and evidence-based clinical diagnostic rules. Through this interaction, DECIS establishes a closed loop of hypothesis formulation, evidence examination, corrective refinement, and report generation, thereby improving both diagnostic accuracy and clinical interpretability.

\subsection{Dual-Evidence Constrained Context}
The goal of Dual-Evidence Constrained Context (DECC) is to construct a diagnostically constrained context for subsequent reasoning and verification. Rather than prompting a LVLM to directly generate a diagnosis from the photograph of the nine cardinal positions of gaze, DECC organizes two complementary sources of evidence into a unified context: visual evidence derived from the input image, and evidence-based clinical diagnostic rules formalized through a doctor-in-the-loop process. In this way, the subsequent reasoning process is constrained not only by visual evidence, but also by evidence-based clinical diagnostic rules as shown in Fig.~\ref{magic_main}.

\subsubsection{Visual Evidence Construction}
\label{sec:svec}
To provide clinically meaningful visual evidence for downstream reasoning and verification, we construct the visual components of DECC through a Complementary Decoupled Attention (CDA) module, as shown in Fig.~\ref{GazeNet}. The motivation is that strabismus diagnosis from the photograph of the nine cardinal positions of gaze is inherently twofold and complementary. In clinical practice, ophthalmologists first inspect alignment-related cues within each gaze position, and then compare abnormal variations across gaze positions to determine the deviation direction and the change in the angle of deviation in different positions of gaze. To better reflect this diagnostic procedure, we explicitly decouple visual modeling into two complementary branches.

Specifically, as shown in Fig.~\ref{GazeNet} the proposed CDA module contains an intra-gaze attention branch and an inter-gaze attention branch. 
The intra-gaze branch is designed to capture local alignment-related cues within a given gaze position, thereby providing visual evidence related to the assessment of ocular misalignment. 
The inter-gaze branch is designed to capture diagnostically relevant variations across gaze positions, which are important for evaluating gaze-dependent changes and analyzing pattern strabismus. 
By separating these two branches, the proposed attention mechanism produces visual evidence that is more consistent with the clinical examination process than a single undifferentiated attention map.

Based on this design, the classifier takes the input photograph of the nine cardinal positions of gaze $\mathbf{X}$ and outputs a confidence-ranked list of candidate subtype predictions together with two complementary forms of attention-based visual evidence. Formally, the candidate prediction set is defined as:
\begin{equation}
\mathcal{F}_{\mathrm{cls}}(\mathbf{X}; \theta_{\mathrm{cls}}) = \{(k_i(\mathbf{X}), p_i(\mathbf{X}))\}_{i=1}^{n}
\label{eq:prediction_output}
\end{equation}
where $\mathcal{F}_{\mathrm{cls}}(\mathbf{X}; \theta_{\mathrm{cls}})$ denotes the classifier parameterized by $\theta_{\mathrm{cls}}$, and $n$ is the number of candidate subtype predictions. For the $i$-th candidate, $k_i(\mathbf{X})$ denotes the predicted subtype and $p_i(\mathbf{X})$ denotes its corresponding confidence score.
In addition, the model produces an intra-gaze heatmap $\mathbf{H}_\mathrm{intra}$ and an inter-gaze heatmap $\mathbf{H}_\mathrm{inter}$, which capture attention patterns within individual gaze positions and across different gaze positions, respectively. In the proposed framework, these heatmaps are used as visual evidence for subsequent evidence-based verification.

Accordingly, the visual-constrained component of DECC is defined as:
\begin{equation}
\mathbf{X}_{\mathrm{visual}} = \{\mathbf{X}, \mathbf{H}_\mathrm{intra}, \mathbf{H}_\mathrm{inter}\}
\label{eq:visual_guidance}
\end{equation}
Together with the confidence-ranked candidate subtype predictions $\{(k_i(\mathbf{X}), p_i(\mathbf{X}))\}_{i=1}^{n}$, these elements form the visual evidence used for subsequent verification under evidence-based clinical diagnostic rules. More results are provided in the supplementary material.

\subsubsection{Evidence-Based Doctor-in-the-Loop Diagnostic Rules Formalization}

\begin{figure*}
\centerline{
\includegraphics[width=1.0\textwidth]{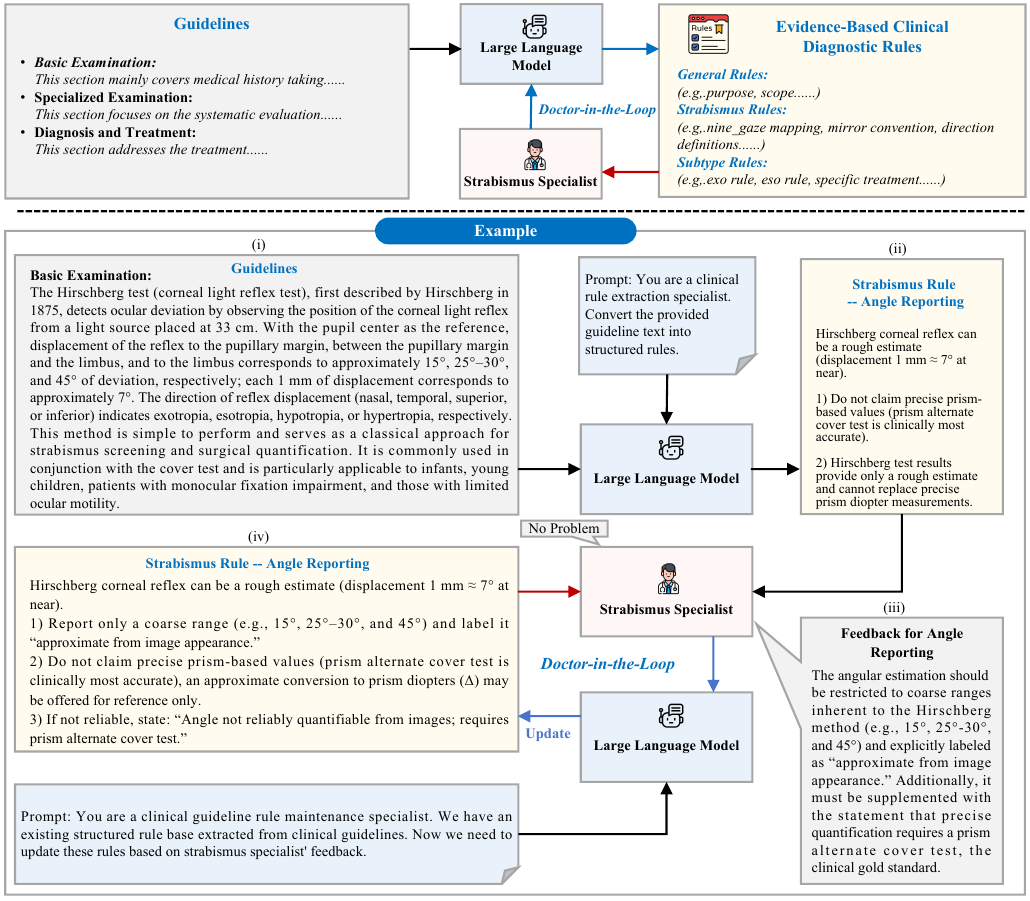}}
\caption{Doctor-in-the-loop evidence-based clinical diagnostic rules formalization. Unstructured clinical guidelines are first converted into candidate diagnostic rules by a large language model (LLM), and then reviewed and refined by strabismus specialists to ensure clinical validity. The bottom panel shows a representative example of this process: (i) clinical knowledge is collected from strabismus guidelines; (ii) the LLM extracts and structures the relevant content into candidate rules; (iii) strabismus specialists review the extracted rules and provide feedback; and (iv) the rule set is iteratively updated through doctor-in-the-loop refinement.
}
\label{knowbase}
\end{figure*}

Visual evidence alone is still insufficient for reliable strabismus diagnosis, because visual evidence must be consistent with clinically relevant diagnostic rules before a diagnostic conclusion can be accepted. To provide this rule-constrained context, we formalize strabismus-related knowledge through the Doctor-in-the-Loop diagnostic rules formalization process. 
The core objective is to convert natural-language clinical texts, such as strabismus-specific guidelines, into structured rule representations that support subsequent evidence review and diagnostic reasoning, as illustrated in Fig.~\ref{knowbase}.

\begin{figure*}
\centerline{
\includegraphics[width=\textwidth]{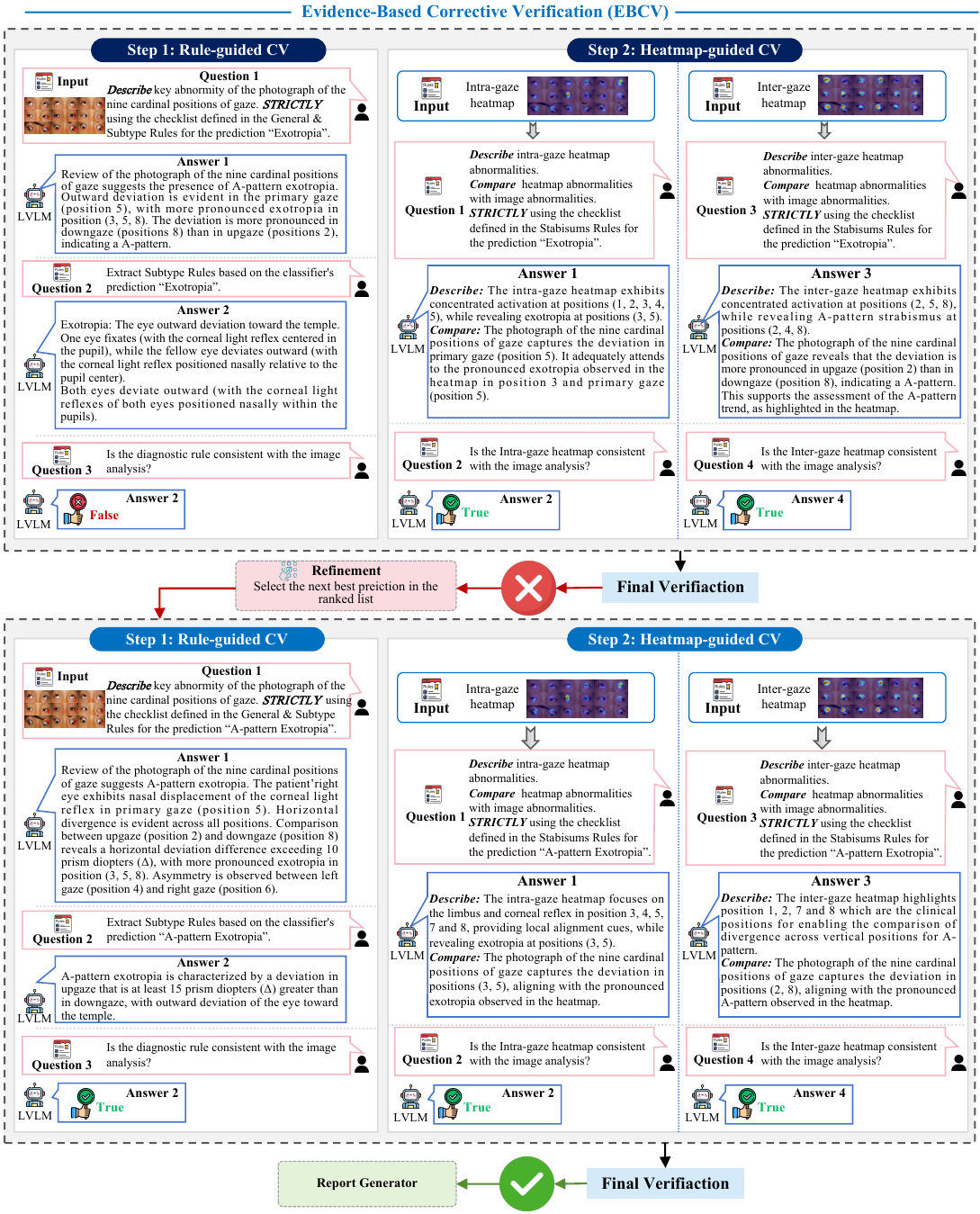}}
\caption{Overview of Evidence-Based Corrective Verification (EBCV). EBCV validates each candidate subtype prediction through two complementary checks: rule-guided CV and heatmap-guided CV. In this example, the current diagnostic hypothesis (Exotropia) fails the rule=-guided CV. Consequently, EBCV rejects this prediction and selects the next-best candidate. The refined prediction, A-pattern Esotropia, passes both CV checks and proceeds to the Generator.
}
\label{ebcv_sample}
\end{figure*}

Specifically, the evidence-based clinical diagnostic rules formalization consists of four steps. 
First, we collect clinically relevant source materials, including official strabismus clinical guidelines, with a primary focus on core evidence-based clinical diagnostic rules, standardized clinical interpretation principles, clinical diagnostic indications, and evidence-referenced treatment thresholds.

These materials provide the initial knowledge base for rule construction.

Second, we prompt an LLM with the collected guidelines and a general extraction instruction to generate an initial rule set. The LLM extracts strabismus-related diagnostic criteria and organizes them into preliminary rules. These rules are then manually reviewed to remove content not directly relevant to strabismus diagnosis, such as epidemiological history, specific patient cases, and general introductory descriptions. The retained rules form the initial rule set, which is subsequently reviewed and refined by specialists.

Third, the extracted rules are reviewed by strabismus specialists in a feedback-driven manner. This stage is designed to ensure clinical agreement of the rule set. In particular, specialists examine whether the extracted rules are consistent with real clinical reasoning, whether ambiguous or potentially misleading statements in the source text have been properly resolved, and whether the terminology has been corrected to conform to professional medical usage. Through this review, clinically inappropriate, oversimplified, or insufficiently specified rule expressions are revised before being used for downstream reasoning.

Finally, the rule set is iteratively refined through doctor-in-the-loop feedback. The LLM updates the structured rules according to the feedback of the specialists, and the revised rules are re-examined until the resulting rule base is clear, clinically meaningful, and suitable for use as diagnostic constraints. 
The final rules are then organized into three levels according to their scope and applicability: 

(1) General Rules $\mathcal{R}_G$: These rules define the overall diagnostic framework, such as the purpose of strabismus examination and the nine cardinal positions of gaze.

(2) Strabismus Rules $\mathcal{R}_S$: These rules establish basic clinical conventions, such as Hirschberg-based measurement principles, deviation direction terminology, and mirror conventions.

(3) Subtype Rules $\mathcal{R}_{sub}^i$: These rules specify diagnostic criteria and treatment-related guidance for the $i$-th strabismus subtype.

These constructed rules constitute the rule-constrained component of DECC for subsequent evidence review and diagnostic reasoning. The entire rule base is then organized into $n$ category-specific entries:
\begin{gather}
\mathcal{R} = \{\mathcal{R}_i\}_{i=1}^{n}
\label{eq:rule_base} \\
\mathcal{R}_i = \{\mathcal{R}_G, \mathcal{R}_S, \mathcal{R}_{sub}^i\}
\label{eq:rule_base1}
\end{gather}

For the $i$-th subtype prediction candidate $k_i(\mathbf{X})$ from the Classifier agent, we retrieve the corresponding rule entry from the evidence-based clinical diagnostic rule base:
\begin{equation}
\begin{split}
\mathcal{R}_{i}(\mathbf{X}) = \mathcal{T}_{\text{retrieve}}(\mathcal{R}, k_i(\mathbf{X}))
\end{split}
\label{eq:rule_retrieval}
\end{equation}
where $\mathcal{T}_{\mathrm{retrieve}}$ denotes the category-indexed rule retrieval function, and $\mathcal{R}_i(\mathbf{X})$ contains the rules corresponding to the predicted strabismus subtype $k_i(\mathbf{X})$.
The retrieved rule component and the visual component $\mathbf{X}_{\mathrm{visual}}$ together form the complete Dual-Evidence Constrained Context:
\begin{equation}
\mathbf{C}_{\mathrm{DECC}} = \{\mathbf{X}_{\mathrm{visual}}, \mathcal{R}_i(\mathbf{X})\}
\label{eq:decc}
\end{equation} 

\subsection{Evidence-Based Corrective Verification}
The purpose of Evidence-Based Corrective Verification (EBCV) is to suppress hallucination by determining whether the current diagnostic hypothesis is supported by the visual evidence of the current case and validated diagnostic rules. In strabismus diagnosis, hallucination occurs when a predicted subtype appears linguistically plausible, yet is not consistent with the visual abnormalities observed in the photograph of the nine cardinal positions of gaze. This support cannot be assessed through a direct comparison between prediction and image, because the former is a diagnostic hypothesis whereas the latter contains only raw visual observations. 

To address this gap, EBCV performs verification through two types of evidence provided by DECC: attention-based visual evidence and structured evidence-based clinical diagnostic rules. 
Specifically, the rules first define the expected abnormalities associated with the current subtype prediction. EBCV then assesses whether these expected abnormalities are consistent with the provided visual evidence, and triggers corrective refinement when inconsistency is detected.
In this way, EBCV transforms hallucination mitigation from constraining the generated response into an explicit and clinically interpretable evidence reviewing process.

\subsubsection{Dual-Evidence Verification}

Given the current candidate subtype prediction $k_i(\mathbf{X})$ and the corresponding Dual-Evidence Constrained Context $\mathbf{C}_{\mathrm{DECC}}$, EBCV performs two complementary consistency checks: rule-guided consistency verification (rule-guided CV) and heatmap-guided consistency verification (heatmap-guided CV), as illustrated in Fig.~\ref{ebcv_sample}.
Rule-guided CV evaluates whether the subtype-specific abnormalities expected by the retrieved diagnostic rules are supported by the case-specific visual evidence. 
Heatmap-guided CV further examines whether the intra-gaze and inter-gaze heatmaps provide compatible attention-based visual evidence. 
These two checks jointly determine whether the current candidate prediction should be accepted or rejected.

The first branch, rule-guided CV, verifies whether the rule-defined expectations for the current candidate prediction are supported by the visual evidence.
For the candidate subtype $k_i(\mathbf{X})$, the corresponding rule entry $\mathcal{R}_i(\mathbf{X})$ specifies the expected abnormalities and gaze-dependent patterns on the photograph of the nine cardinal positions of gaze.
The LVLM is then required to describe the observed visual abnormalities, and the Verifier agent compares these two types of abnormalities. 
This rule-guided CV process is illustrated by Step 1 in the Fig.~\ref{ebcv_sample}.
This process evaluates the consistency between clinical rules and visual evidence for the current candidate prediction.

The second branch, heatmap-guided CV, verifies whether the abnormal regions highlighted by the model’s attention maps are consistent with those identified from the original image. 
The intra-gaze and inter-gaze heatmaps are evaluated separately, as illustrauted by Step 1 in the Fig.~\ref{ebcv_sample}. 
For each heatmap, the LVLM describes the attended abnormal regions and compares them with the visual evidence observed in the photograph of the nine cardinal positions of gaze during rule-guided CV. 
A heatmap-guided CV result is considered positive when the attended regions align with the observed abnormalities; otherwise, it is considered negative, indicating that the model may have relied on irrelevant or misleading regions.

The final verification outcome is obtained by combining the rule-guided and heatmap-guided consistency verification. 
Rule-guided CV is treated as a necessary condition because it evaluates the clinical validity of the current candidate prediction. Specifically, a candidate subtype should not be accepted unless the abnormalities required by its corresponding diagnostic rules are supported by the visual evidence. 
Heatmap-guided CV provides auxiliary confirmation of the model’s attention behavior. The intra-gaze and inter-gaze heatmaps capture complementary forms of attention-based visual evidence: the former reflects local ocular abnormalities within individual gaze positions, while the latter reflects gaze-dependent variations across positions.
Since different strabismus subtypes may rely more strongly on either local ocular misalignment (\eg, Esotropia and Exotropia) or cross-gaze variation (\eg, A-pattern and V-pattern), requiring both heatmap-guided checks to pass would be overly restrictive. Therefore, the heatmap-guided condition is satisfied when at least one of the two heatmaps is consistent with the observed ocular abnormalities.

\subsubsection{Corrective Refinement}
When the current candidate prediction fails dual-evidence verification, EBCV rejects it rather than accepting the generated conclusion. Such rejection indicates that the predicted subtype lacks sufficient support from case-specific visual evidence under the corresponding diagnostic rules. This rejection mechanism suppresses hallucination by preventing unsupported predictions from being treated as valid diagnostic conclusions.

If the current candidate prediction fails dual-evidence verification, DECIS selects the next candidate from the confidence-ranked list and reconstructs the corresponding DECC for another round of verification. Formally, the candidate index at iteration $t$ is updated as:
\begin{equation}
r_t = 
\begin{cases} 
1, & t = 1 \\
r_{t-1} + 1, & t > 1
\end{cases}
\label{eq:candidate_update_v2}
\end{equation}
where $r_t$ denotes the index of the candidate prediction evaluated at iteration $t$.
The DECC is reconstructed as:
\begin{equation}
\mathbf{C}_{\mathrm{DECC}} = \{\mathbf{X}_{\mathrm{visual}}, \mathcal{R}_{r_{t}}(\mathbf{X})\}
\label{eq:decc_update}
\end{equation}

Importantly, the failed verification outcome is used to reject the unsupported candidate prediction and evaluate the next candidate under the same evidence review protocol. By iteratively filtering out predictions that are not supported by visual evidence or diagnostic rules, DECIS reduces hallucinated diagnoses and moves toward an evidence-grounded final decision.
Once a candidate prediction passes EBCV, the verified results are passed to the Generator agent for report generation.

\subsection{Report Generation}
This stage is designed not merely to generate a fluent textual summary, but to organize the verified diagnostic outcome into a clinically interpretable and evidence-traceable form. In this way, the report generation serves as the final stage of the systematic reasoning loop established by DECIS, converting evidence-grounded verification results into a report format that is suitable for clinical reading and review.

Specifically, the generated report includes the key abnormalities, strabismus pattern and angle description, consistency verification results, treatment-related recommendations, and the corresponding rules used during reasoning. 
Each component is grounded in the verified findings produced by the preceding stages. 
The primary diagnosis summarizes the abnormalities and pattern and angle descriptions derived from visual evidence, the corrective verification generated from EBCV, and the treatment recommendations generated under the associated evidence-based clinical diagnostic rules. Therefore, the report is not a free-form response generated directly from the image, but a structured synthesis of diagnosis, visual evidence, verification results, and evidence-based clinical diagnostic rules.
Further details regarding the LVLM prompts are provided in the supplementary material.

\section{Experiments and Results}
\subsection{Experimental Setup}
\subsubsection{Dataset}
Existing strabismus datasets are often limited in scale and lack fine-grained subtype annotations, which restricts their utility for evaluating clinically meaningful diagnosis. To address this issue, we constructed a fine-grained strabismus benchmark comprising clinical data from 1,075 patients collected at Joint Shantou International Eye Center of Shantou University and the Chinese University of Hong Kong. This study was approved by the Ethics Committee of Joint Shantou International Eye Center of Shantou University and the Chinese University of Hong Kong (approval no. EC 20260325(2)-P06, and registered at the Chinese Clinical Trial Registry (ChiCTR, registration number: ChiCTR2600122191). A waiver of informed consent was granted due to the retrospective cohort of deidentified images captured for training purposes.
For each patient, the dataset includes a photograph of the nine cardinal positions of gaze, annotated by senior ophthalmologists with fine-grained subtype labels. The subtype annotations include six categories: Exotropia, A-pattern Exotropia, V-pattern Exotropia, Esotropia, V-pattern Esotropia, and Vertical strabismus.

Following standard practice in medical image analysis, the dataset was divided into training, validation, and test sets at a ratio of 6:3:1. This benchmark supports both subtype classification and diagnostic evaluation in our experiments. More detailed dataset information is provided in the supplementary material.

\subsubsection{Comparison Methods}
For the diagnostic evaluation, we considered two group of methods: prompt-based methods and fine-tuned medical LVLMs. The prompt-based methods include Reflexion \cite{shinn2023reflexion}, CoVe \cite{dhuliawala2024chain}, while the fine-tuned medical LVLM include HealthGPT \cite{lin2025healthgpt} and HuatuoGPT-Vision \cite{chen2024towards}. All methods were evaluated using the same input images and task instructions.

For strabismus subtype classification, we compared DECIS with representative image classification models, including VGG-16 \cite{simonyan2014very}, ResNet-50 \cite{he2016deep}, DenseNet \cite{huang2017densely}, ViT \cite{dosovitskiy2020image}, Swin Transformer \cite{liu2021swin}, DeiT \cite{touvron2021training}, and Vision Mamba \cite{zhu2024vision}. In addition, we included CI-GNN \cite{zheng2025causality}, a recent method specifically developed for strabismus subtype classification, as a task-specific method.

\subsubsection{Implementation Details}
The overall experimental pipeline consists of two stages: strabismus subtype classification and diagnosis. In the classification stage, the Classifier agent was implemented in PyTorch and trained on a single NVIDIA GeForce RTX 4090 GPU. 
Input images were resized to $192 \times 576$ to preserve the spatial layout of the photograph of the nine cardinal positions of gaze, and were then partitioned into non-overlapping $16 \times 16$ patches. Optimization was performed using SGD with momentum of 0.9, weight decay of $5 \times 10^{-5}$, and an initial learning rate of $1 \times 10^{-3}$ with cosine annealing. The model was trained for 500 epochs with standard data augmentation, including random horizontal flipping and color jittering.

In the diagnosis stage, the Verifier and Generator agent were instantiated using three representative LVLMs: Gemini-3-Flash-Preview \cite{google2026gemini3flashpreview}, GPT-5.2 \cite{openai2025gpt52systemcard}, and Qwen3-VL-Plus \cite{bai2025qwen3}.

\subsection{Evaluation Metrics}
To comprehensively evaluate the proposed framework, we assess both the clinical reliability of generated diagnostic reports and the performance of strabismus subtype classification.

\subsubsection{Clinical Reliability Metrics}
Following the human evaluation protocol of PhraseAug \cite{PhraseAug}, we evaluate the quality of the generated diagnostic reports along three dimensions: Clinical Consistency (Consistency), Visual Alignment (Alignment), and Contextual Completeness (Completeness). Five strabismus specialists independently rated each report on a 5-point Likert scale and were blinded to the method identity.

Consistency assesses whether the report agrees with accepted clinical knowledge and diagnostic reasoning. 
Alignment evaluates whether the main diagnostic conclusions are supported by corresponding visual evidence. 
Completeness measures whether the report covers the key aspects required for strabismus assessment, including abnormality description, gaze-dependent variation, subtype-related pattern analysis, and angle approximation.
The three metrics jointly measure the clinical acceptability, evidence support, and practical usefulness of the generated reports. The final score for each metric was obtained by averaging ratings across specialists and test cases.

\subsubsection{Classification Performance Metrics}
To evaluate strabismus subtype classification performance, we report two groups of metrics: task-specific accuracy metrics and weighted joint-classification metrics. $\text{Acc}_{\text{Dir}}$ and $\text{Acc}_{\text{Ang}}$ are used to evaluate the two subtasks separately.  
Specifically, $\text{Acc}_{\text{Dir}}$ evaluates the recognition of deviation direction, (\ie, esotropia, exotropia, and vertical strabismus), while $\text{Acc}_{\text{Ang}}$ evaluates the recognition of angle variations across gaze positions (\ie, A-pattern, V-pattern, and absence of A- or V-pattern):
\begin{gather}
    \text{Acc}_{\text{Dir}} = \frac{1}{N} \sum_{i=1}^{N}
    \mathrm{I}\bigl(\hat{y}_i^{\text{Dir}} = y_i^{\text{Dir}}\bigr),\\
    \text{Acc}_{\text{Ang}} = \frac{1}{N} \sum_{i=1}^{N}
    \mathrm{I}\bigl(\hat{y}_i^{\text{Ang}} = y_i^{\text{Ang}}\bigr),
\end{gather}
where $y_i$ and $\hat{y}_i$ denote the ground-truth and predicted labels of the $i$-th sample, $N$ is the number of test samples, and $\mathrm{I}(\cdot)$ is the indicator function. Here, the superscripts $\text{Dir}$ and $\text{Ang}$ correspond to deviation direction and angle variation, respectively.

In addition to task-specific accuracy, we report weighted precision ($\text{W-P}$), weighted recall ($\text{W-R}$), 
and weighted F1-score ($\text{W-F1}$) to evaluate the overall performance of joint subtype classification. 
Each sample is assigned to one of six joint classes formed by the combination of 
deviation direction and angle variation. The detailed mathematical definitions of these metrics are provided in the supplementary.

\subsection{Main Results}

\begin{table*}[t]
    \centering
    \footnotesize
    \setlength{\tabcolsep}{3pt}
    \caption{Comparison of clinical reliability and classification performance across diagnostic methods. Clinical reliability metrics (Consistency, Alignment, Completeness) are rated on a 5-point Likert scale (5 = highest). Classification metrics ($\text{Acc}_{\text{Dir}}$, $\text{Acc}_{\text{Ang}}$, weighted precision (W-P), weighted recall (W-R), weighted F1 (W-F1)) are reported as percentages [\%]. All methods, including prompt-based approaches (Reflexion, CoVe) and fine-tuned medical LVLMs (HealthGPT, HuatuoGPT-Vision), were evaluated under identical input conditions and task instructions.}
    \resizebox{\textwidth}{!}{%
    \begin{tabular}{l|c|c|c|c|c|c|c|c}
        \toprule
        \textbf{Methods} &
        \textbf{Consistency } &
        \textbf{Alignment } &
        \textbf{Completeness } &
        $\textbf{Acc}_\textbf{Dir}$ &
        $\textbf{Acc}_\textbf{Ang}$ &
        \textbf{W-P } &
        \textbf{W-R } &
        \textbf{W-F1 } \\
        \midrule 
        \midrule

        \multicolumn{9}{l}{\textit{\textbf{Prompt-based Methods}}} \\
        \midrule
        Qwen + CoVe\cite{dhuliawala2024chain} & 1.85 & 1.98 & 1.79 & 63.6 & 23.4 & 32.4 & 17.8 & 11.1 \\
        Qwen + Reflexion\cite{shinn2023reflexion} & 1.44 & 1.35 & 1.37 & 43.9 & 40.2 & 42.3 & 15.0 & 13.6 \\
        GPT + CoVe\cite{dhuliawala2024chain} & 1.84 & 1.87 & 1.80 & 60.7 & 45.8 & 43.1 & 26.2 & 30.7 \\
        GPT + Reflexion\cite{shinn2023reflexion} & 2.03 & 2.00 & 1.98 & 70.1 & 41.1 & 74.9 & 38.3 & 41.5 \\
        Gemini-3-flash-preview + CoVe\cite{dhuliawala2024chain} & 2.43 & 2.33 & 2.45 & 71.0 & 18.7 & 79.1 & 16.8 & 9.6 \\
        Gemini-3-flash-preview + Reflexion\cite{shinn2023reflexion} & 2.12 & 1.99 & 2.06 & 65.4 & 16.8 & 2.3 & 15.0 & 4.0 \\
        \midrule 

        \multicolumn{9}{l}{\textit{\textbf{Fine-tuning-based Models}}} \\
        \midrule
        HuatuoGPT-Vision\cite{chen2024towards} & 2.16 & 2.08 & 2.05 & 56.1 & 64.5 & 36.5 & 31.8 & 33.9 \\
        HealthGPT\cite{lin2025healthgpt} & 2.15 & 1.99 & 2.05 & 51.4 & 70.1 & 42.8 & 35.5 & 38.8 \\
        \midrule 
        \midrule

        \multicolumn{9}{l}{\textit{\textbf{Ours}}} \\
        \midrule
        Qwen + DECIS & 3.42 & 3.35 & 4.20 & 94.4 & 88.8 & 83.3 & 83.2 & 79.3 \\
        GPT + DECIS & 3.64 & 3.95 & 3.90 & 93.5 & 93.5 & 84.5 & 87.9 & 85.2 \\
        \textbf{Gemini-3-flash-preview + DECIS} & \textbf{4.52} & \textbf{4.14} & \textbf{4.41} & \textbf{95.3} & \textbf{97.2} & \textbf{91.1} & \textbf{92.5} & \textbf{91.3} \\
        \bottomrule
    \end{tabular}%
    }
    \label{ebm_mainresult}
\end{table*}

\begin{table*}[t]
    \centering
    \footnotesize
    \setlength{\tabcolsep}{3pt}
    \caption{Ablation study of DECIS components. We evaluate the contributions of three key components: the diagnostic rules and visual evidence in DECC, and the consistency verification mechanism in EBCV. Clinical reliability metrics (Consistency, Alignment, Completeness) are rated on a 5-point Likert scale (5 = highest), and classification metrics ($\text{Acc}_{\text{Dir}}$, $\text{Acc}_{\text{Ang}}$, weighted precision (W-P), weighted recall (W-R), weighted F1 (W-F1)) are reported as percentages [\%]. All results are obtained using Gemini-3-Flash-Preview as the default LVLM.}
    \resizebox{\linewidth}{!}{%
    \begin{tabular}{cccccccccccc}
        \toprule
        \textbf{Setting} &
        \textbf{Diagnostic Rules} &
        \textbf{Visual Evidence} &
        \textbf{Consistency Verification} &
        \textbf{Consistency} &
        \textbf{Alignment} &
        \textbf{Completeness} &
        $\textbf{Acc}_\textbf{Dir}$  &
        $\textbf{Acc}_\textbf{Ang}$  &
        \textbf{W-P } &
        \textbf{W-R } &
        \textbf{W-F1 } \\
        \midrule 

        (a) &  &  &  & 1.04 & 1.05 & 1.02 & 92.5 & 82.2 & 58.1 & 75.7 & 65.7 \\
        (b) & \cmark &  &  & 3.15 & 3.16 & 3.64 & 92.5 & 82.2 & 58.1 & 75.7 & 65.7\\
        (c) &  & \cmark &  & 2.19 & 2.24 & 2.16 & 92.5 & 82.2 & 58.1 & 75.7 & 65.7\\
        (d) &  &  & \cmark & 3.37 & 3.76 & 3.42 & 93.5 & 87.9 & 77.4 & 81.3 & 79.0 \\
        (e) & \cmark & \cmark &  & 3.90 & 3.96 & 4.00 & 92.5 & 82.2 & 58.1 & 75.7 & 65.7 \\
        (f) & \cmark &  & \cmark & 3.97 & 4.03 & 4.19 & 92.5 & 97.2 & 86.6 & 90.7 & 88.4 \\
        (g) &  & \cmark & \cmark & 3.38 & 3.56 & 3.59 & 95.3 & 95.3 & 89.6 & 90.7 & 89.7 \\
        \textbf{(h)} & \textbf{\cmark} & \textbf{\cmark} & \textbf{\cmark} &
        \textbf{4.52} & \textbf{4.14} & \textbf{4.41} & \textbf{95.3} & \textbf{97.2} & \textbf{91.1} & \textbf{92.5} & \textbf{91.3} \\
        \bottomrule
    \end{tabular}%
    }
    \label{ebm_ablation}
\end{table*}

\begin{table}[t]
    \centering
    \footnotesize
    \setlength{\tabcolsep}{3pt}
    \caption{Performance comparison with representative classification methods. Classification metrics ($\text{Acc}_{\text{Dir}}$, $\text{Acc}_{\text{Ang}}$, weighted precision (W-P), weighted recall (W-R), weighted F1 (W-F1)) are reported as percentages [\%].}
    \resizebox{\linewidth}{!}{%
    \begin{tabular}{lccccc}
        \toprule
        \textbf{Model} &
        $\textbf{Acc}_\textbf{Dir}$  &
        $\textbf{Acc}_\textbf{Ang}$  &
        \textbf{W-P } &
        \textbf{W-R } &
        \textbf{W-F1 } \\
        \midrule

        VGG-16~\cite{simonyan2014very} 
        & 89.7 & 77.6 & 63.2 & 68.2 & 65.1 \\

        ResNet-50~\cite{he2016deep} 
        & 93.5 & 81.3 & 69.7 & 74.8 & 68.1 \\

        DenseNet~\cite{huang2017densely} 
        & 91.6 & 81.3 & 62.2 & 75.7 & 68.2 \\

        ViT~\cite{dosovitskiy2020image} 
        & 88.8 & 82.2 & 55.9 & 72.9 & 63.3 \\

        Swin Transformer~\cite{liu2021swin} 
        & 90.7 & 78.5 & 55.9 & 70.1 & 62.2 \\

        DeiT~\cite{touvron2021training} 
        & 73.8 & 81.3 & 44.3 & 57.0 & 48.9 \\

        Vision Mamba~\cite{zhu2024vision} 
        & 76.6 & 81.3 & 48.2 & 60.8 & 52.0 \\

        CI-GNN~\cite{zheng2025causality} 
        & 90.5 & 86.9 & 75.5 & 77.6 & 72.0 \\

        \midrule
        \textbf{Our DECIS} 
        & \textbf{95.3} & \textbf{97.2} & \textbf{91.1} & \textbf{92.5} & \textbf{91.3} \\
        \bottomrule
    \end{tabular}%
    }
    \label{performance}
\end{table}

\subsubsection{Clinical Reliability}
As shown in Fig.~\ref{ebm_mainresult}, we first evaluate the clinical reliability of generated diagnostic reports and the ability to suppress hallucination.
Prompt-based methods, including Reflexion and CoVe, and fine-tuned medical LVLMs, including HealthGPT and HuatuoGPT-Vision, obtain relatively low scores in Consistency, Alignment, and Completeness.
In contrast, DECIS consistently achieves higher scores across different LVLM backbones, suggesting that the proposed evidence-based consistency verification process effectively reduces linguistically plausible but unsupported outputs. 
In particular, DECIS with Gemini-3-Flash-Preview achieves the best performance on all three metrics. 
These results demonstrate that DECIS substantially improves the clinical reliability by grounding report generation in visual evidence and evidence-based clinical diagnostic rules.

\subsubsection{Strabismus Subtype Classification Performance}
As shown in Fig.~\ref{ebm_mainresult}, we further evaluate the effect of DECIS on strabismus subtype classification. 
DECIS consistently achieves the best results across all classification metrics. 
Among different LVLM backbones, DECIS with Gemini-3-Flash-Preview obtains the highest overall performance, with 95.3\% $\text{Acc}_{\text{Dir}}$, 97.2\% $\text{Acc}_{\text{Deg}}$, 91.1\% weighted precision, 92.5\% weighted recall, and 91.3\% weighted F1 score. 
The improvement in $\text{Acc}_{\text{Deg}}$ suggests that modeling inter-gaze variation is particularly useful for recognizing A- and V-pattern strabismus. 
In addition, Fig.~\ref{performance} shows that DECIS significantly outperforms representative classification methods, including VGG, ResNet, DenseNet, ViT, Swin Transformer, DeiT, Vision Mamba, and the task-specific CI-GNN method.

\begin{figure*}[t]

	\centerline{\includegraphics[width=\textwidth]{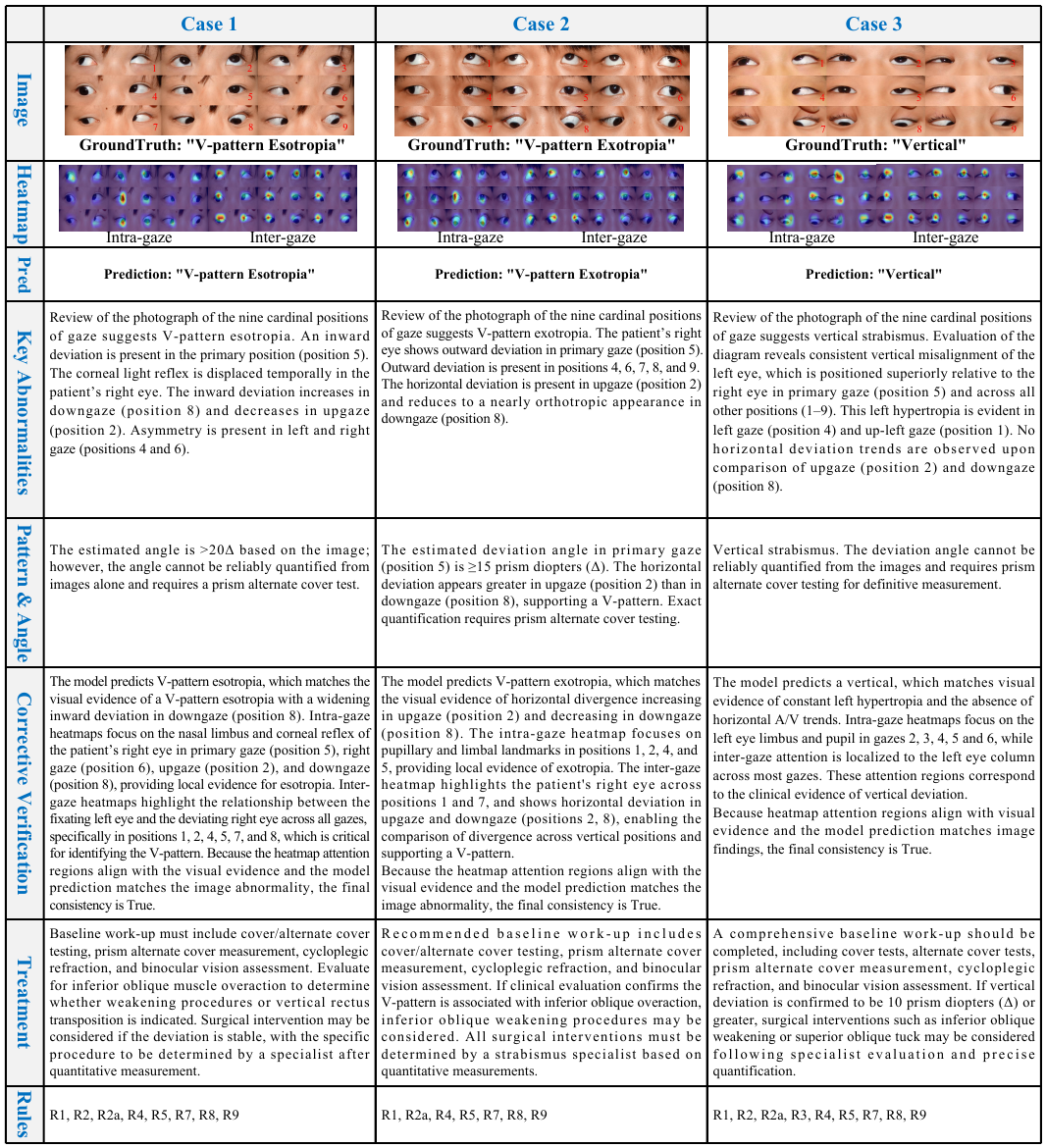}}
	\caption{Representative diagnostic reports generated by DECIS using Gemini-3-Flash-Preview as the LVLM backbone. Each report comprises several components: prediction, key abnormalities, pattern and angle analysis, corrective verification, treatment recommendations, and rules used.}
	\label{ebmgemini}

\end{figure*}

\subsection{Ablation Study}

We conduct ablation studies using Gemini-3-Flash-Preview as the default LVLM to evaluate the contributions of the proposed components (\ie, evidence-based clinical diagnostic rules (diagnostic rules) and visual evidence in DECC, and consistency verification mechanism in EBCV). 
As shown in Fig.~\ref{ebm_ablation}, the model performs poorly when all three components are removed, achieving only 1.04, 1.05, and 1.02 in Consistency, Alignment, and Completeness, respectively. This indicates that direct generation without evidence constraints is insufficient for clinically reliable strabismus diagnosis.

Adding diagnostic rules alone substantially improves the clinical reliability of generated reports, especially Completeness, which increases from 1.02 to 3.64. This suggests that rule-based clinical knowledge helps the model organize diagnostically relevant content. Adding visual evidence alone also improves the results, but its effect is weaker than diagnostic rules in this setting. When diagnostic rules and visual evidence are combined, the scores further increase to 3.90, 3.96, and 4.00, showing that the two components of DECC provide complementary support for report generation.

The consistency verification mechanism brings additional gains, especially in subtype classification. Compared with using diagnostic rules and visual evidence without verification, adding consistency verification improves weighted precision from 58.1\% to 91.1\%, weighted recall from 75.7\% to 92.5\%, and weighted F1 score from 65.7\% to 91.3\%. 
This shows that DECC alone can improve the quality of generated reports, but explicit verification is necessary to reject unsupported candidate predictions and select more reliable candidates. 
The full DECIS configuration achieves the best overall performance across all metrics, demonstrating that DECC and EBCV jointly contribute to subtype classification, hallucination suppression, and clinical reliability.

\subsection{Qualitative Evaluation of Diagnostic Report}

To further evaluate the interpretability of DECIS, we qualitatively analyze representative diagnostic reports generated by the proposed framework, as shown in Fig.~\ref{ebmgemini}. The visualization includes three representative cases covering V-pattern Esotropia, V-pattern Exotropia, and Vertical strabismus. For each case, DECIS presents the original photograph of the nine cardinal positions of gaze, subtype prediction, the intra-gaze and inter-gaze heatmaps, key abnormalities, pattern and angle analysis, consistency verification, treatment-related recommendations, and the diagnostic rules used. This structured layout allows the diagnostic conclusion to be traced back to both visual evidence and diagnostic rules.

In Case 1, DECIS identifies V-pattern Esotropia. The report describes inward deviation in the primary gaze and a stronger horizontal deviation in downgaze than in upgaze, which supports the V-pattern diagnosis. The intra-gaze heatmap highlights the patient’s right eye in primary gaze, which corresponds to the local ocular misalignment observed in the original image. 
In Case 2, DECIS identifies V-pattern Exotropia by describing horizontal divergence across gaze positions and a larger deviation in upgaze than in downgaze. The heatmaps highlight gaze positions that are clinically relevant for comparing horizontal deviation across vertical gaze directions. 
In Case 3, DECIS identifies Vertical strabismus by describing consistent vertical misalignment of the left eye across gaze positions, with heatmap attention localized to regions corresponding to vertical deviation.

These examples show that DECIS does not generate free-form diagnostic descriptions directly from images. Instead, it organizes the verified results into a clinically reviewable report, where the final prediction is supported by key abnormalities, pattern and angle analysis, consistency verification, and the corresponding rules used during diagnosis. Overall, the qualitative results demonstrate that DECIS can produce evidence-traceable diagnostic reports that make the diagnostic process more transparent and clinically interpretable.

\section{Conclusion}
In this work, we proposed DECIS, an evidence-based multi-agent framework for interpretable strabismus diagnosis. DECIS addresses the limited transparency of conventional deep learning methods and the hallucination risk of LVLM-based approaches by reformulating strabismus diagnosis as an evidence-constrained and verifiable process. Specifically, candidate predictions are examined against visual evidence and evidence-based clinical diagnostic rules rules before the final diagnosis is generated. By integrating Dual-Evidence Constrained Context and Evidence-Based Corrective Verification, DECIS improves fine-grained subtype prediction while producing diagnostic reports with stronger clinical interpretability and trustworthiness. Extensive experiments show that DECIS provides an effective solution for evidence-supported and clinically reliable strabismus diagnosis.

\section{Limitations and Future Work}
DECIS is currently evaluated on a single strabismus subtype diagnosis benchmark. Further validation with larger cohorts and multi-center datasets is needed to assess its robustness across different imaging protocols, acquisition conditions, and patient populations. Beyond this task, DECIS provides an extensible framework for evidence-supported medical diagnosis. By replacing the visual evidence module and evidence-based clinical diagnostic rules with task-specific counterparts, it can be adapted to broader medical fields, such as radiology, dermatology, pathology, and endoscopy.
Future work will evaluate DECIS in these domains to further validate its generalizability.

\section*{References}
\bibliographystyle{IEEEtran}

\bibliography{reference}

\end{document}